\documentclass[10pt,journal,compsoc]{IEEEtran}
\newif\ifpeerreview

\peerreviewfalse

\usepackage[compress]{cite}
\usepackage{url}
\usepackage{amsmath,amssymb,graphicx}
\usepackage{xcolor}
\usepackage{makecell}
\usepackage{booktabs}
\usepackage{multirow}

\usepackage{lipsum} 
\usepackage[switch]{lineno}

\makeatletter
\newcommand{\supplementtitle}{%
\twocolumn[{%
\begin{@twocolumnfalse}
\begin{center}
    {\LARGE Supplementary Material\par}
    \vspace{0.75em}

    {\LARGE Null-Space Diffusion Distillation Unlocks Speed, Fidelity and Realism in Lensless Imaging\par}
    \vspace{1em}

\end{center}
\vspace{1.5em}
\end{@twocolumnfalse}
}]%
}
\makeatother

\newcommand{\red}[1]{\textcolor{black}{#1}}

\usepackage{pifont}
\newcommand{\cmark}{\textcolor{green!60!black}{\ding{51}}} 
\newcommand{\xmark}{\textcolor{red!70!black}{\ding{55}}}   

\newcommand{\paperID}{18}

\title{Null-Space Diffusion Distillation Unlocks Speed, Fidelity and Realism in Lensless Imaging}

\author{Jose~Reinaldo Cunha Santos A V Silva Neto, Hodaka Kawachi, Yasushi Yagi, 
        and~Tomoya Nakamura
\IEEEcompsocitemizethanks{\IEEEcompsocthanksitem Jose Reinaldo Cunha Santos A V Silva Neto is with the D3 Center, The University of Osaka, Osaka,
Japan.\protect\\
E-mail: vieira@is.ids.osaka-u.ac.jp
\IEEEcompsocthanksitem H. Kawachi is with SANKEN, The University of Osaka.
\IEEEcompsocthanksitem Y. Yagi is with D3 center, The University of Osaka.
\IEEEcompsocthanksitem T. Nakamura is with Grad. Sch. of Eng. Sci., The University of Osaka.}
}

\begin{document}

\IEEEtitleabstractindextext{%
\begin{abstract}
Lensless imaging reconstructs scenes from highly multiplexed measurements, resulting in a severely ill-posed inverse problem. In this work, we identify a fundamental trade-off between measurement consistency, perceptual quality, and inference speed across lensless reconstruction paradigms. Traditional methods favor consistency but produce perceptually degraded results, supervised approaches achieve high-quality reconstructions with fast inference but may violate physical constraints, and diffusion-prior methods achieve high perceptual quality and consistency---particularly when structured constraints such as range--null decomposition are used---but remain slow due to iterative sampling.
Motivated by this observation, we propose \emph{Null-Space Diffusion Distillation} (NSDD), a single-pass reconstruction model that distills structured diffusion-prior inference into an efficient feed-forward network. NSDD learns to produce high-quality reconstructions that preserve measurement consistency while avoiding costly iterative sampling.
Experimental results demonstrate that NSDD achieves \red{perceptual quality and consistency competitive with diffusion-prior methods, while providing significantly faster inference and offering a favorable balance across all three objectives.} \red{Furthermore, ablation experiments show that distilling the range--null decomposition improves reconstruction quality and robustness over unstructured full-reconstruction distillation, including on unseen real scenes.} These results highlight the potential of structure-aware distillation for efficient lensless imaging. Code is available at github.com/JRCSAVSN/NullSpaceDiffusionDistillation.

\end{abstract}

}

\maketitle

\IEEEraisesectionheading{
  \section{Introduction}\label{sec:introduction}
}

\begin{figure*}
    \centering
    \includegraphics[width=0.9\textwidth]{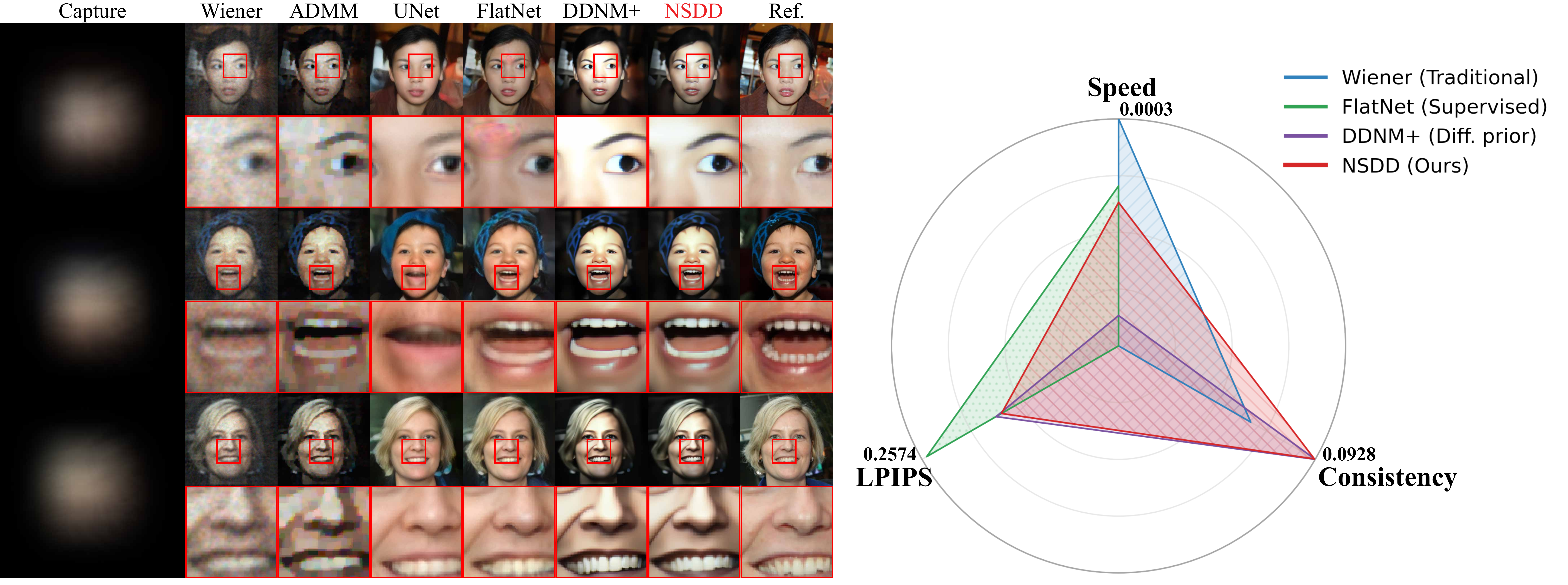}
    \caption{\textbf{Overview of reconstruction trade-offs in lensless imaging.} 
\textbf{Left:} Representative reconstructions from traditional, supervised, diffusion-prior optimization, and our (NSDD) method, compared to the reference image. While supervised methods enable fast inference with high quality and classical methods enforce measurement consistency, NSDD achieves both. 
\textbf{Right:} Trade-off between measurement consistency, inference time \red{and LPIPS perceptual quality} across reconstruction \red{paradigms}, showing that NSDD \red{achieves a balance between these three challenging objectives. Axes are reversed, meaning that higher (i.e., outward) scores are better and speed axis is presented in log scale for ease of visualization.}}
    \label{fig:1_teaser}
\end{figure*}

Lensless cameras encode visual information through coded optical elements that multiplex incoming light onto a sensor, replacing conventional lens-based image formation. While this enables compact and flexible imaging systems~\cite{Xu_etal_lenslessrobot_IEEETrob_2025, Hua_etal_SphericalCamera_arxiv_2023}, it results in a highly ill-posed inverse problem: the captured measurement is a spatially entangled representation of the scene, further corrupted by noise and model mismatch. In this setting, practical lensless imaging requires balancing three competing objectives: \emph{measurement consistency}, to ensure fidelity to the physical imaging model; \emph{perceptual quality}, to produce visually realistic reconstructions; and \emph{inference speed}, to enable practical deployment.

Reconstruction methods for lensless imaging have evolved along multiple directions. Early approaches rely on physics-driven optimization, such as Wiener deconvolution or iterative methods~\cite{Boyd_etal_ADMM_2011_TrendsML, Bioucas_etal_TwIST_2007_TransImgProc, Beck_Teboulle_FISTA_2009_ICASSP, Monakhova_etal_UDN_2021_OptExp}, which explicitly enforce measurement consistency but often struggle to produce visually realistic results. More recent work has shifted toward supervised learning-based approaches, where neural networks are trained to map lensless measurements to reference images, achieving strong perceptual quality with fast inference~\cite{Monakhova_etal_leADMM_2019_Optica, Kingshott_etal_PrimalDualNet_2022_Optica, Khan_etal_flatnet_PAMI_2022, Cai_etal_phocolens_2024_NeurIPS, Yosef_Giryes_difuzcam_2025_SciRep, Li_etal_MWDNs_2023_OptExpress}. However, these methods rely on paired training targets that are typically not generated by the same forward model as the lensless measurements—for example, using beam-splitter setups with lensed cameras or display-based captures—which can introduce biases and limit reconstruction quality/fidelity.

More recently, diffusion-prior-based methods have emerged as a powerful alternative in general inverse imaging, achieving strong performance in tasks such as inpainting, colorization, and motion deblurring~\cite{Chung_etal_DPS_2023_ICLR, wang_etal_ddnm_2023_iclr, Song_etal_piGD_2023_ICLR, Kim__etal_flowdps_2025_ICCV}. These approaches combine expressive pretrained generative priors with explicit measurement constraints during inference, enabling reconstructions that are both perceptually realistic and physically consistent. Despite this success, their adoption in lensless imaging has remained limited. Moreover, their reliance on iterative sampling procedures results in slow inference time, which hinders practical deployment.

In this work, we show that lensless imaging reconstruction is governed by a fundamental trade-off between three competing objectives: measurement consistency, perceptual quality, and inference speed. Traditional methods favor consistency by explicitly enforcing the forward model, but often produce perceptually degraded results. Supervised approaches achieve high perceptual quality with fast inference, but may not fully respect the physical constraints of the imaging system. Diffusion-prior-based methods achieve strong perceptual quality with improved fidelity, at the cost of high inference time due to iterative sampling. We analyze this trade-off in the lensless setting and show that these paradigms indeed occupy distinct regions in this space, as illustrated in Fig.~\ref{fig:1_teaser}. A more detailed explanation is shown in Sec.~\ref{sec:supervised}. 

Motivated by this observation, we propose \emph{Null-Space Diffusion Distillation} (NSDD), a single-pass reconstruction model that distills the structured, noise-aware behavior of a diffusion-based inverse solver into a fast single-pass feed-forward network. By conditioning the network on the same control signals used by to guide the diffusion-prior solver, NSDD learns a residual correction that approximates the inference process of a structured diffusion-based reconstruction approach, improving reconstruction speed while retaining both perceptual quality and measurement consistency. In this way, NSDD retains the benefits of diffusion-based inference while eliminating the need for costly iterative sampling. Sec.~\ref{sec:nsdd} provides a more detailed introduction and analysis of our method.

The contributions of this work are as follows:
\begin{itemize}
    \item \textbf{Analysis of lensless reconstruction trade-offs.} We provide the first empirical analysis of reconstruction paradigms in lensless imaging, showing that traditional, supervised, and diffusion-prior methods occupy distinct points in a three-way trade-off between measurement consistency, perceptual quality, and inference speed.
    
    \item \textbf{Diffusion priors for lensless imaging.} We show that diffusion-prior-based iterative methods—previously studied in general inverse problems—are effective in the lensless setting, achieving a favorable balance between perceptual quality and measurement fidelity.
    
    \item \textbf{Structure-aware distillation (NSDD).} Motivated by the previous analyses, we propose Null-Space Diffusion Distillation, a single-pass model that distills a structured diffusion-based solver into a fast reconstruction network, achieving high perceptual quality, strong measurement consistency, and \red{faster inference speed}.
    
    \item \textbf{Qualitative validation on real lensless captures.} Our experiments are done on real lensless imaging systems (Lensless-FFHQ and PhlatCam), thus supporting our claims using real systems.
\end{itemize}

\section{Imaging model and generative priors}
\label{sec:background}

\begin{table*}[t]
\vspace{-1.5mm}
\centering
\caption{\textbf{Overview of inverse imaging reconstruction strategies.}
Unsupervised ground-truth-free methods rely on explicit priors and measurement consistency, while paired-target supervised approaches learn a direct mapping from measurements to images. $K$ denotes solver iterations or unrolled stages, and $T$ the number of neural function evaluations.}
\setlength{\tabcolsep}{2.8pt}
\begin{tabular}{l c c c c}
\toprule
Method & 
Ground-truth-free & 
Inference Family & 
Inference cost & 
Data-independent $A$/$A^\dagger$ \\
\midrule
\makecell{\textbf{Ground-truth-free (unsupervised)}} \\
Wiener &  \cmark & Closed-form / linear & 1 & \cmark \\
ADMM~\cite{Boyd_etal_ADMM_2011_TrendsML} &  \cmark & Iterative optimization & $K$ iters & \cmark \\
DPS~\cite{Chung_etal_DPS_2023_ICLR} &  \cmark & Diffusion iterative & $T$ NFEs & \cmark \\
DDNM+~\cite{wang_etal_ddnm_2023_iclr} &  \cmark & Diffusion iterative & $T$ NFEs & \cmark \\
\midrule
\multicolumn{5}{l}{\textbf{Supervised}} \\
LeADMM~\cite{Monakhova_etal_leADMM_2019_Optica} &  \xmark & Unrolled ADMM + UNet\cite{Ronnenberger_etal_unet_2015_miccai} & $K$ stages & \cmark \\
FlatNet~\cite{Khan_etal_flatnet_PAMI_2022} &  \xmark & Inversion + UNet & 1 & \xmark \\
PhoCoLens~\cite{Cai_etal_phocolens_2024_NeurIPS} &  \xmark & Inversion + Diffusion & $T$ NFEs & \xmark \\
PrimalDualNet~\cite{Kingshott_etal_PrimalDualNet_2022_Optica} &  \xmark & Unrolled primal-dual & $K$ stages & \cmark \\
\bottomrule
\end{tabular}
\label{tab:method_characteristics}
\vspace{-2mm}
\end{table*}

\subsection{Lensless imaging}

Many computational imaging systems can be modeled as linear inverse problems of the form
\begin{equation}
    \boldsymbol{y}=\boldsymbol{A}\boldsymbol{x}+\boldsymbol{n},
\end{equation}
where $\boldsymbol{y}\in\mathbb{R}^{1\times N_y}$ denotes the measurement, $\boldsymbol{x}\in\mathbb{R}^{1\times N_x}$ the unknown signal of interest, $\boldsymbol{A}\in\mathbb{R}^{N_y\times N_x}$ the forward operator, and $\boldsymbol{n}$ measurement noise. In many practical settings, including lensless imaging, the operator $\boldsymbol{A}$ is ill-conditioned or non-invertible due to the multiplexing of information, making the recovery of $\boldsymbol{x}$ from $\boldsymbol{y}$ a highly ill-posed problem.


In lensless cameras, the lenses are replaced by coded masks that multiplex the incoming light onto the sensor, resulting in a spatially entangled measurement. Under the assumption of a shift-invariant point spread function (PSF), the forward model can be approximated as a two dimensional convolution~\cite{Boominathan_etal_LenslessCameraReview_Optica_2022},
\begin{equation}
    \boldsymbol{y} = \boldsymbol{v}*\boldsymbol{h}+\boldsymbol{n},
\end{equation}
where $\boldsymbol{y}\in\mathbb{R}^{H\times W}$ and $\boldsymbol{v}\in\mathbb{R}^{H\times W}$ are the unflattened measurement and latent image, respectively, $\boldsymbol{h}\in\mathbb{R}^{H\times W}$ is the calibrated PSF, and $*$ denotes 2D convolution.

Reconstruction in lensless imaging is therefore commonly formulated as a regularized inverse problem~\cite{Boominathan_etal_LenslessCameraReview_Optica_2022, Li_etal_LenslessReview_FundResearch_2024},
\begin{equation}
    \tilde{\boldsymbol{x}} =  
    \operatorname*{argmin}_{\boldsymbol{x}\geq 0}\Big\{
    \| \boldsymbol{y} - \boldsymbol{A}\boldsymbol{x} \|^2_2 + \tau \Psi(\boldsymbol{x})\Big\},
    \label{eq:trad_opt}
\end{equation}
where $\Psi(\cdot)$ is a regularization term, such as total variation~\cite{Rudin_etal_2dtv_PhysD_1992}, that encodes prior assumptions on natural images. This formulation highlights the need to balance measurement consistency with prior information to resolve the inherent ambiguity of the inverse problem.

\subsection{Priors in lensless imaging}

The regularized formulation in Eq.~(\ref{eq:trad_opt}) highlights that reconstruction in inverse problems relies on combining measurement consistency with prior information on the unknown signal. Early approaches typically employ hand-crafted priors, such as sparsity or total variation, which promote simple structural properties in the reconstructed image. These methods are commonly solved using closed-form Wiener filtering or iterative optimization algorithms such as ADMM and provide reconstructions that are consistent with the measurements but often producing noisy lower quality reconstructions.

Recent advances have explored replacing hand-crafted priors with learned models. In particular, supervised approaches train neural networks to directly map measurements $\boldsymbol{y}$ to high-quality images $\boldsymbol{x}$ using paired training data. These methods have demonstrated strong empirical performance and are widely adopted for photorealistic reconstruction in lensless imaging. However, since the mapping is learned from data rather than explicitly constrained by the forward model, measurement consistency is not necessarily enforced during inference. More generally, these approaches can be interpreted as implicitly learning a prior over natural images through supervised training. While effective in many cases, their reliance on paired targets and implicit enforcement of the imaging model motivates alternative formulations that more directly incorporate both prior information and measurement constraints. We summarize the main characteristics of representative reconstruction approaches in Table~\ref{tab:method_characteristics}, highlighting differences in training requirements, inference strategies, computational cost, and data-dependence for model calibration (i.e., $A$ and $A^\dagger$).

\subsection{Diffusion priors for inverse problems}

Diffusion models provide a powerful class of generative priors by learning to progressively denoise samples from a noise distribution. Starting from a noisy input $\boldsymbol{x}_t$, a neural network $S_{\theta}(\boldsymbol{x}_t, t)$ is trained to approximate the score function $\nabla_{\boldsymbol{x}_t} \log p_t(\boldsymbol{x}_t)$, which guides the reconstruction toward the data distribution. In practice, this enables computing a denoised estimate of the underlying signal,
\begin{equation}
    \hat{\boldsymbol{x}}_0 \approx \red{\frac{1}{\sqrt{\bar{\alpha}_t}}}\bigg(\boldsymbol{x}_t + (1 - \bar{\alpha}_t) S_{\theta}(\boldsymbol{x}_t, t)\bigg),
\end{equation}
where $\hat{\boldsymbol{x}}_0$ represents an estimate of the clean image at timestep $t$. By iteratively refining $\boldsymbol{x}_t$, diffusion models can generate high-quality samples and serve as expressive image priors.

To solve inverse problems, diffusion models are combined with measurement constraints during the sampling process. A common strategy is to guide the diffusion using the likelihood of the observations. In Diffusion Posterior Sampling (DPS)~\cite{Chung_etal_DPS_2023_ICLR}, this is achieved by augmenting the score function with a gradient derived from the measurement consistency objective,
\begin{equation}
    \nabla_{\boldsymbol{x}_t} \log p(\boldsymbol{x}_t|\boldsymbol{y}) \approx \nabla_{\boldsymbol{x}_t} \log p(\boldsymbol{x}_t) - \zeta \nabla_{\boldsymbol{x}_t} \|\boldsymbol{y} - \boldsymbol{A}(\hat{\boldsymbol{x}}_0)\|_2^2,
\end{equation}
where $\hat{\boldsymbol{x}}_0$ denotes the denoised estimate. While effective, such approaches operate in the measurement space and may struggle to maintain a stable balance between perceptual quality and measurement consistency.

An alternative strategy is to enforce measurement consistency directly in the reconstruction space. Methods such as DDNM~\cite{wang_etal_ddnm_2023_iclr} leverage the range--null decomposition of the forward operator to separate the reconstruction into a measurement-consistent component and a perceptual component. The denoised estimate is updated as
\begin{equation}
    \hat{\boldsymbol{x}}'_0 = \boldsymbol{A}^{\dagger} \boldsymbol{y} + (\boldsymbol{I} - \boldsymbol{A}^{\dagger} \boldsymbol{A})\hat{\boldsymbol{x}}_0,
\end{equation}
where $\boldsymbol{A}^{\dagger} \boldsymbol{y}$ enforces consistency in the range space, and $(\boldsymbol{I} - \boldsymbol{A}^{\dagger} \boldsymbol{A})\hat{\boldsymbol{x}}_0$ applies diffusion-based denoising in the null space. 

To account for noisy measurements, DDNM$+$ introduces a relaxed, noise-aware correction that replaces the hard projection with a weighted update toward measurement consistency. This improves robustness while preserving the underlying range--null decomposition structure. These approaches provide a principled way to combine diffusion priors with the physical constraints of the imaging system.

\begin{table*}[t]
\centering
\scriptsize
\setlength{\tabcolsep}{3.5pt}
\renewcommand{\arraystretch}{1.15}

\caption{\textbf{Quantitative comparison.}
\textbf{x-FR}: full-reference image-domain metric.
\textbf{x-NR}: no-reference perceptual metric.
\textbf{y-space}: measurement-domain consistency.
\textbf{Bold} and \underline{underlined} indicate \textbf{best} and \underline{second best}, respectively. Metrics are highlighted separately for unsupervised/supervised methods.}

\resizebox{\textwidth}{!}{
\begin{tabular}{l c c c c c c c c c}
\toprule
\multirow{3}{*}{Method}
& \multicolumn{5}{c}{\textbf{Lensless-FFHQ}}
& \multicolumn{4}{c}{\textbf{PhlatCam}} \\
\cmidrule(lr){2-6} \cmidrule(lr){7-10}
&
\makecell{\textbf{Time} $\downarrow$ \\ \scriptsize (s)}
& \makecell{\textbf{Consist.} $\downarrow$ \\ \scriptsize $\|\mathbf{y}-A\hat{\mathbf{x}}\|$ $(\times 10^{-4})$}
& \makecell{\textbf{SSIM} $\uparrow$ \\ \scriptsize x-FR}
& \makecell{\textbf{LPIPS} $\downarrow$ \\ \scriptsize x-FR}
& \makecell{\textbf{CLIP-IQA} $\uparrow$ \\ \scriptsize x-NR}
& \makecell{\textbf{Consist.} $\downarrow$ \\ \scriptsize $\|\mathbf{y}-A\hat{\mathbf{x}}\|$ $(\times 10^{-10})$}
& \makecell{\textbf{SSIM} $\uparrow$ \\ \scriptsize x-FR}
& \makecell{\textbf{LPIPS} $\downarrow$ \\ \scriptsize x-FR}
& \makecell{\textbf{CLIP-IQA} $\uparrow$ \\ \scriptsize x-NR}
\\
\midrule

\multicolumn{10}{l}{\textbf{Unsupervised}} \\
Wiener
& \textbf{0.0003} & 1.918          & \textbf{0.5432} & 0.5738          & 0.2041
                  & 1.0721         & 0.4141          & 0.6043          & 0.1576 \\

ADMM~\cite{Boyd_etal_ADMM_2011_TrendsML}
& 0.1471          & 0.1595         & 0.4877          & 0.4435          & 0.3787
                  & 0.6235         & 0.4031          & 0.6033          & 0.0737 \\

DPS~\cite{Chung_etal_DPS_2023_ICLR}
& 116.4288        & 0.1408         & 0.2803          & 0.4323          & \textbf{0.5869}
                  & 0.4422         & 0.2088          & 0.7352          & \textbf{0.7052} \\

DDNM+~\cite{wang_etal_ddnm_2023_iclr}
& 20.8908         & \textbf{0.0928} & 0.4830                & \textbf{0.3717} & \underline{0.5180}
                  & \textbf{0.3268} & \underline{0.4462}    & \textbf{0.5245} & \underline{0.2777} \\

NSDD (ours)
& \underline{0.0342} & \underline{0.1156} & \underline{0.4984} & \underline{0.3807} & 0.4954
                     & \underline{0.4128} & \textbf{0.4464}    & \underline{0.5633} & 0.2380 \\

\midrule
\multicolumn{10}{l}{\textbf{Supervised}} \\

UNet~\cite{Monakhova_etal_leADMM_2019_Optica}
& 0.0132 & 5.269          & 0.4985          & 0.2922          & 0.5038
                  & 1.9989         & 0.3917          & 0.5661          & 0.4648 \\

FlatNet~\cite{Khan_etal_flatnet_PAMI_2022}
& 0.0134          & 5.660           & 0.5042          & 0.2574          & 0.5496
                  & 1.9134          & 0.4474          & 0.4929          & 0.5988 \\

LeADMM-U~\cite{Monakhova_etal_leADMM_2019_Optica}
& 0.0606          & 5.487          & 0.5043          & 0.2497          & 0.5533
                  & 2.0479         & \textbf{0.4498} & 0.4658          & 0.5097 \\

PrimalDualNet~\cite{Kingshott_etal_PrimalDualNet_2022_Optica}
& 0.0292          & \textbf{5.115} & 0.4143          & 0.4074          & 0.4451
                  & 2.2606         & 0.2530          & 0.6202          & 0.2734 \\

MultiWienerNet~\cite{Yanny_etal_MultiWienerNet_2022_optica}
& \textbf{0.0118} & 5.554          & \textbf{0.5098}          & 0.2356          & 0.5370
                  & 2.1277         & 0.4450          & 0.4952          & 0.4799 \\

PhoCoLens~\cite{Cai_etal_phocolens_2024_NeurIPS}
& 118.3432        & 5.706           & 0.4752          & \textbf{0.2242} & \textbf{0.7169}
                  & \textbf{1.8959} & 0.3832          & \textbf{0.4622} & \textbf{0.7717} \\
\bottomrule
\end{tabular}
}
\label{tab:restoration_metrics}
\end{table*}


 \subsection{Distillation for diffusion models}

A major limitation of diffusion models is their high computational cost, as high-quality samples typically require many iterative denoising steps. To address this, a large body of work has explored \emph{distillation} techniques that approximate a multi-step diffusion process with a few-step or single-step model while preserving generation quality.

Existing approaches differ in how the student model is trained. \emph{Step-distillation} progressively reduces the number of sampling steps while retaining fidelity~\cite{Salimans_Ho_DiffDist1_ICLR_2022, Meng_etal_DiffDist1_CVPR_2023}. \emph{Consistency-based} methods learn a mapping that is consistent across noise levels, enabling direct few-step generation~\cite{Song_etal_DiffDist2_ICML_2023, Luo_etal_DiffDist2_arxiv_2023, Luo_etal_DiffDist3_arxiv_2023}. \emph{Distribution-matching} approaches fit a single-step generator to the teacher distribution using approximate score or KL objectives~\cite{Yin_etal_DiffDist4_CVPR_2024, Yin_etal_DiffDist4_NIPS_2024}. Finally, \emph{adversarial} distillation combines teacher supervision with adversarial training to stabilize low-step sampling~\cite{Sauer_etal_AdvDiffDist_ECCV_2024}.

While effective for unconditional generation, these approaches are typically designed to match the output distribution of the teacher and do not explicitly account for measurement constraints arising in inverse problems. This motivates distillation strategies that preserve the structure of physically consistent inference procedures, rather than only their output distribution.

In this work, we build on this perspective and propose to distill a diffusion-based inverse solver while approximating its range--null decomposition strategy. Specifically, we adopt an offline distillation strategy inspired by~\cite{nimrod_etal_KoopmanDistill_neurips_2025}, where the student is conditioned on measurement-dependent control signals (i.e., $\boldsymbol{y}$ and $\boldsymbol{A}^\dagger \boldsymbol{y}$) and trained to reproduce single-pass reconstructions from a DDNM$+$ teacher.

\section{Analysis of Reconstruction Strategies for Lensless Imaging}
\label{sec:supervised}

In this section, we analyze the three reconstruction paradigms (i.e., traditional, supervised, and diffusion-prior methods) and show that they occupy distinct points in a trade-off between perceptual quality, measurement consistency, and inference speed. Traditional methods favor consistency, supervised methods favor perceptual quality with fast inference, and diffusion-based approaches enable high quality and consistent images but at the cost of more costly computational complexity.

\subsection{Experimental setup}
\label{sec:experimental_setup}


\noindent\textbf{Dataset.}
We evaluate reconstruction performance on the Lensless-FFHQ~\cite{Neto_etal_SelfNeuralLensless_2025_OptReview} \red{and PhlatCam~\cite{Khan_etal_flatnet_PAMI_2022}} datasets. \red{Lensless-FFHQ} consists of $21{,}000$ lensless captures obtained using a prototype camera with an amplitude radial coded mask~\cite{Neto_etal_RadialOpt_2023_IeeeTci}. Each measurement corresponds to an image from the FFHQ dataset~\cite{karras_etal_ffhq_2021_TPAMI}, enabling comparison against high-quality reference images. \red{Additional information about train-test splits for Lensless-FFHQ and PhlatCam datasets are presented in the supplementary materials. The pretrained diffusion model on FFHQ uses the same train-test split as the Lensless-FFHQ, to ensure a fair comparison.}

\noindent\textbf{Reconstruction methods.}
We compare representative approaches from three reconstruction paradigms: traditional methods (Wiener, ADMM), supervised methods (UNet, FlatNet, leADMM, PrimalDualNet, \red{MultiWienerNet and PhoCoLens}), and diffusion-prior-based methods (DPS, DDNM$+$). For diffusion-\red{prior}-based approaches, we use a pretrained model trained on FFHQ dataset following Chung et al.~\cite{Chung_etal_DPS_2023_ICLR} \red{for the Lensless-FFHQ experiments, and pretrained on ImageNet~\cite{Deng_etal_ImageNet_2009_CVPR} for PhlatCam experiments,} with a predefined $1,000$ diffusion steps. Consistent with prior work~\cite{Cai_etal_phocolens_2024_NeurIPS}, the Wiener filter is used as an approximation of the pseudo-inverse operation in DDNM$+$. All methods are evaluated under the same forward model and measurement conditions.

\noindent\textbf{Experimental setting.}
We perform a multifaceted evaluation of representative methods of all three reconstruction classes (i.e., traditional, supervised, diffusion-prior), that aims to quantify these methods' reconstruction quality, measurement consistency and inference speed. We use $200$ test samples to calculate the reconstruction metrics (e.g., LPIPS, SSIM), and compare representative methods of all three reconstruction categories: traditional methods (Wiener, ADMM), supervised methods (UNet, FlatNet), and diffusion-based iterative methods (DPS, DDNM+). 

\noindent\textbf{Evaluation metrics.}
\red{We evaluate each method using complementary metrics that capture both reconstruction quality and physical fidelity. Image-domain reconstruction quality is measured using SSIM for structural similarity, LPIPS~\cite{zhang_etal_lpips_2018_CVPR} for perceptual similarity, and CLIP-IQA~\cite{Wang_etal_ClipIQA_2023_AAAI} for no-reference perceptual quality. Measurement-domain fidelity is quantified by the consistency error, computed as the mean-squared error between the measured lensless image $\mathbf{y}$ and the predicted measurement $\mathbf{A}\hat{\mathbf{x}}$. We additionally report inference time for each method.}

\red{For image-domain metrics, all reconstructions from unsupervised and distilled methods are first normalized to the range $[0,1]$ to ensure compatibility with the reference images and provide a fair comparison against supervised methods, whose outputs are trained directly against these references. We then apply the same restoration-metric preprocessing pipeline to all evaluated reconstructions: images are cropped to the valid imaging region, geometrically aligned to the reference using an affine transformation estimated via feature matching, and intensity-normalized to match the reference dynamic range before computing SSIM, LPIPS, and CLIP-IQA.}

\red{For measurement consistency, we account for the fact that the normalized reconstructions may differ from the physical measurement scale. Therefore, before computing the consistency error, we estimate per-image scale and offset parameters $a$ and $b$ by least squares to minimize $|\mathbf{y}-\mathbf{A}(a\hat{\mathbf{x}}+b)|_2^2$. The consistency metric is then computed after this scale-bias compensation. Since no single metric captures both perceptual realism and physical fidelity, we analyze image-domain quality, measurement consistency, and inference time jointly. Full details of the preprocessing pipeline, sensor-to-diffusion-domain mapping through cropping and padding, dataset-specific train/test splits, and evaluation protocols are provided in the supplementary materials.}

\noindent\textbf{Computational environment.} All supervised training experiments are performed on \red{either} a Quadro RTX 8000 \red{or an NVIDIA A100 GPU}. However, inference for all methods is done on an RTX 5090 GPU on an Ubuntu 24.04 OS with Intel i9-10900K, and the per-image latency is measured as wall-clock inference time excluding data I/O and averaged over 200 images.




\subsection{Trade-off between consistency, perceptual quality, and inference speed}


Fig.~\ref{fig:1_teaser}-(right) illustrates the trade-off between measurement consistency, perceptual quality, and inference time across reconstruction methods. Tab.~\ref{tab:restoration_metrics} reports the corresponding quantitative values for all evaluation metrics, while Fig.~\ref{fig:lensless_recon} provides representative qualitative reconstructions for each method. Together, these results show that different approaches occupy distinct regions in this metrics space, revealing a fundamental trade-off between physical fidelity, perceptual realism, and computational cost.

Traditional methods achieve strong measurement consistency but produce perceptually degraded reconstructions. In contrast, supervised methods show low perceptual error but often exhibit significant inconsistency with the measurements. Diffusion-prior-based methods provide a more balanced compromise, improving perceptual quality while maintaining reasonable consistency; however, this comes at the expense of substantially higher inference time due to their iterative nature.

These observations indicate that existing reconstruction approaches do not simultaneously optimize all three objectives, motivating the need for methods that can better achieve perceptual quality, measurement consistency, and inference speed. \red{PrimalDualNet underperforms on PhlatCam, likely because its back-projection-based layers require tuning for the $950\times950$ preprocessed measurements, while we used the same architecture as for Lensless-FFHQ.}

\subsection{Behavior of diffusion-based methods}

Although diffusion-based approaches improve the trade-off between perceptual quality and measurement consistency, their behavior depends strongly on how measurement constraints are enforced. Methods based on likelihood-score guidance, such as DPS, operate in the measurement space and may exhibit unstable behavior when balancing these objectives. In contrast, methods such as DDNM$+$ explicitly separate the reconstruction into range-space and null-space components, enforcing measurement consistency through projection while leveraging the diffusion prior to refine perceptual details. This structured approach leads to more stable and reliable reconstructions.

These observations suggest that the manner in which measurement constraints are incorporated into the diffusion process guidance plays a critical role in reconstruction performance.A parameter-sweep analysis (see supplementary) shows that DPS fails to produce physically consistent and perceptually plausible reconstructions across a wide range of hyperparameters, whereas DDNM+ consistently reconstructs the correct scene structure and exhibits a trade-off between noise and oversmoothing controlled by its noise parameter.

\subsection{Discussion}

The above analysis highlights a fundamental trade-off in lensless reconstruction between measurement consistency, perceptual quality, and inference speed. Physics-driven and supervised methods occupy opposite ends of this spectrum, while diffusion-based approaches provide a promising middle ground.

Among diffusion-based methods, approaches that explicitly leverage the range--null decomposition of the forward model exhibit more stable behavior and better adherence to the measurement constraints. However, their reliance on iterative sampling results in high computational cost.

These observations motivate the development of methods that retain the structured, physically consistent behavior of diffusion-based inference while significantly reducing computational complexity. 
\section{Null-space diffusion distillation for lensless imaging}
\label{sec:nsdd}

\begin{figure*}
    \centering
    \includegraphics[width=0.9\linewidth]{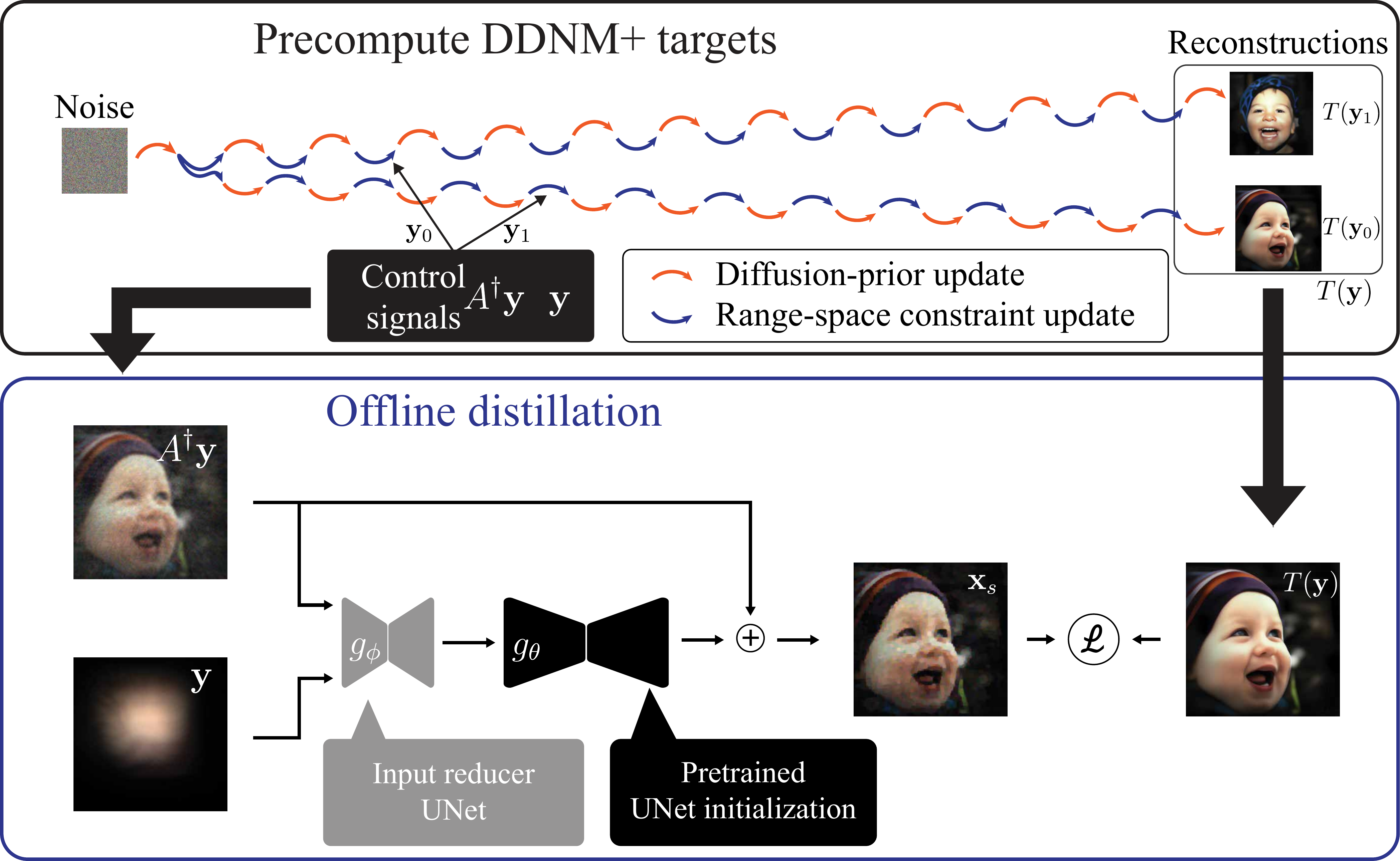}
    \caption{NSDD overview. Top: Precomputation of the DDNM$+$ targets \(T(\mathbf y)\) using a fixed seed ($\textbf{x}_T$). Bottom: Distill the concatenate measurement \(\mathbf y\) with the range anchor \(\mathbf A^\dagger\mathbf y\) along channel dimension, reduce 6\(\to\)3 channels via a small UNet \(g_\phi\), and predict a residual correction with the pretrained diffusion UNet \(g_\theta\). The output \(\hat{\mathbf x}_s=\mathbf A^\dagger\mathbf y+\tilde{\mathbf z}\) is a single-pass reconstruction.}
    \label{fig:nsdd_diagram}
\end{figure*}

\begin{figure*}
    \centering
    \includegraphics[width=\linewidth]{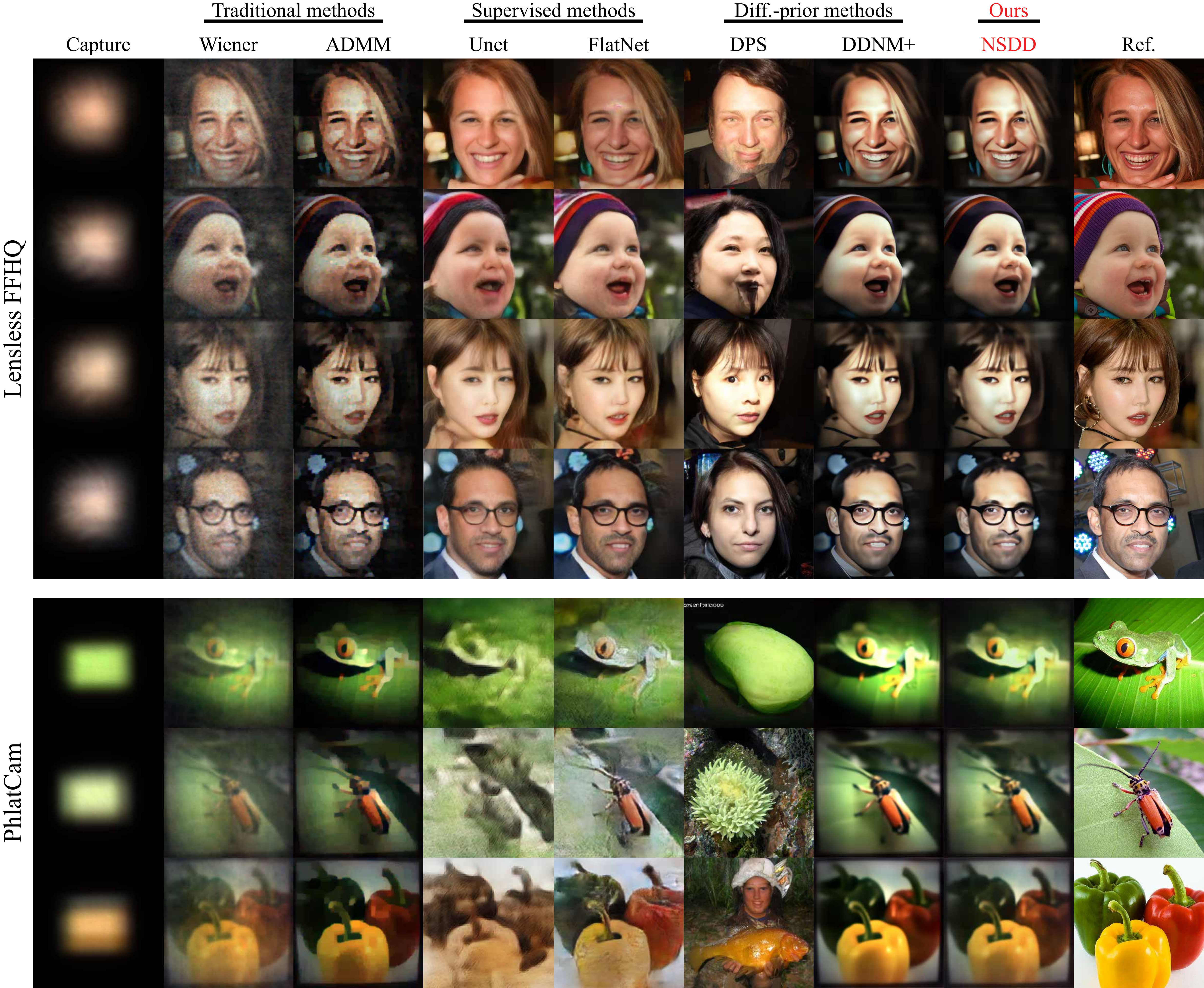}
    \caption{Qualitative reconstructions for Lensless-FFHQ and PhlatCam datasets. Left to right: lensless measurement (capture), Wiener, ADMM, Unet, FlatNet, DPS, DDNM$+$ (teacher), NSDD (ours), reference image (Ref.).}
    \label{fig:lensless_recon}
\end{figure*}

\subsection{Motivation}

Building on the analysis in Section~3, we observe that diffusion-based methods provide a favorable balance between perceptual quality and measurement consistency compared to traditional and supervised approaches. In particular, structured formulations such as DDNM$+$, which explicitly leverage the range--null decomposition of the forward model, yield stable and physically consistent reconstructions.

However, these methods rely on iterative sampling procedures that incur high computational cost, limiting their practical applicability. This motivates the development of models that retain the structured behavior of diffusion-based inference while significantly reducing inference time.

To this end, we propose \emph{Null-Space Diffusion Distillation} (NSDD), a single-pass reconstruction model that distills a DDNM$+$ solver while approximating its range--null structure. Distillation is performed offline by precomputing teacher reconstructions, enabling training without paired ground-truth images.

\subsection{Method}

\noindent\textbf{Teacher targets (offline).}
For each training measurement \(\mathbf y\), we run DDNM$+$ with a \emph{fixed} random seed (same \(\mathbf{x}_T\)) to obtain a deterministic teacher reconstruction \(T(\mathbf y)\). Although diffusion sampling is stochastic, fixing the initialization yields stable targets and simplifies student training without explicit conditioning on the input noise.

\noindent\textbf{Physics-aware conditioning.}
To approximate the structured inference process of DDNM$+$, we condition the student on both the measurement and a physics-based anchor:
\[
\mathbf z = \mathrm{concat}\big(\mathbf y,\; \mathbf A^\dagger \mathbf y\big)\in\mathbb{R}^{H\times W\times 6}.
\]
Here, \(\mathbf y\in\mathbb{R}^{H\times W\times 3}\) is the lensless measurement and \(\mathbf A^\dagger\mathbf y\in\mathbb{R}^{H\times W\times 3}\) is its pseudo-inverse reconstruction. The anchor \(\mathbf A^\dagger\mathbf y\) provides an estimate of the measurement-consistent structure (range-space), while the measurement \(\mathbf y\) retains information about the physical constraints and noise characteristics of the inverse problem. We perform an ablation study to evaluate the impact of each conditioning signal separately in Sec.~\ref{sec:nsdd_ablation}.

\noindent\textbf{Student predictor.}
We reuse a pretrained diffusion UNet \(g_\theta\) as the main backbone and fine-tune it to predict a residual correction relative to the anchor \(\mathbf A^\dagger\mathbf y\), similar to DDNM+. Since \(g_\theta\) expects 3-channel inputs, we prepend a lightweight input reducer \(g_\phi\) UNet (architecture details in the supplementary) that maps the 6-channel tensor \(\mathbf z\) to 3 channels:
\[
\tilde{\mathbf z}=g_\phi(\mathbf z)\in\mathbb{R}^{H\times W\times 3}, \qquad
\hat{\mathbf r}=g_\theta(\tilde{\mathbf z}).
\]
The final reconstruction is then written as
\begin{equation}
\hat{\mathbf x}_s \;=\; \mathbf A^\dagger \mathbf y \;+\; \hat{\mathbf r}.
\label{eq:nsdd_compose}
\end{equation}
Unlike the noiseless DDNM formulation, DDNM$+$ does not enforce a strict decomposition in which only the null space is updated. Instead, its noise-aware correction allows limited adjustment of the range-space component to improve robustness under noisy measurements. Accordingly, the residual \(\hat{\mathbf r}\) learned by NSDD should be interpreted as a teacher-consistent correction around the anchor \(\mathbf A^\dagger\mathbf y\), capturing both null-space refinements and denoising behavior inherited from DDNM$+$.

\noindent\textbf{Loss.}
We train the student using a mean squared error objective between the predicted reconstruction \(\hat{\mathbf x}_s\) and the teacher output \(T(\mathbf y)\):
\begin{equation}
\mathcal L_{\text{MSE}} \;=\; \frac{1}{N}\sum_{i=1}^{N}\big\|\,\hat{\mathbf x}_s^{(i)}-T(\mathbf y_i)\,\big\|_2^2.
\label{eq:nsdd_loss}
\end{equation}
This objective transfers the reconstruction behavior of the teacher without requiring paired ground-truth images or explicit physics-based loss terms. In practice, the anchor \(\mathbf A^\dagger\mathbf y\) stabilizes training by providing a physically meaningful initialization, while the residual branch learns the corrections needed to approximate the noisy DDNM$+$ solver in a single forward pass.

\subsection{Implementation details}

\noindent\textbf{Training setup.}
We train NSDD on both Lensless-FFHQ and PhlatCam datasets. In both cases, teacher reconstructions are generated offline using DDNM$+$ and cached for efficient training. Additional information about train-test splits and diffusion models are presented in the supplementary material.

\noindent\textbf{Pretrained models.}
We use the diffusion model implementation of guided diffusion~\cite{dhariwal_nichol_guideddiff_NIPS_2021}. For Lensless-FFHQ, we use the model pretrained on FFHQ following Chung et al.~\cite{Chung_etal_DPS_2023_ICLR}. For PhlatCam, we use an unconditional diffusion model pretrained on ImageNet~\cite{dhariwal_nichol_guideddiff_NIPS_2021}, since the scene content is more diverse and not restricted to faces. In both cases, the student backbone \(g_\theta\) is initialized from the corresponding pretrained diffusion UNet and fine-tuned jointly with the input reducer \(g_\phi\).

\noindent\textbf{Teacher configuration.}
The DDNM$+$ teacher uses a fixed noise seed and a noise parameter \(\sigma_y=0.6\), selected to balance smoothing and noise in the reconstruction. The pseudo-inverse \(\mathbf A^\dagger\) is implemented using a Wiener filter and shared between teacher and student to avoid operator mismatch.

\noindent\textbf{Architecture and optimization.}
The input reducer \(g_\phi\) is a compact UNet, while \(g_\theta\) is the pretrained diffusion UNet adapted for residual prediction with a fixed timestep. Both networks are trained jointly using Adam with a learning rate of \(10^{-4}\) for 100 epochs. Specific details of the input reducer network are presented in the supplementary material.

\noindent\textbf{Evaluation.}
We evaluate NSDD using the same preprocessing, alignment, and normalization pipeline described in Section~3, ensuring consistent comparison across all methods.

\subsection{Discussion}

Table~\ref{tab:restoration_metrics} reports quantitative results on Lensless-FFHQ, where NSDD achieves a favorable trade-off between perceptual quality, measurement consistency, and inference time. While closely matching the perceptual quality of diffusion-based methods such as DDNM$+$, NSDD operates in a single forward pass, resulting in \red{significantly faster inference when compared to the teacher DDNM$+$}. Compared to traditional methods, NSDD significantly improves perceptual quality, and compared to supervised methods, it maintains stronger adherence to the measurement constraints.

Figure~\ref{fig:lensless_recon} presents qualitative reconstructions for both Lensless-FFHQ and PhlatCam datasets. On Lensless-FFHQ, NSDD produces reconstructions that are visually consistent with the teacher while avoiding hallucinations observed in alternative diffusion-based approaches (i.e., DPS), and artifacts observed in supervised methods (see also Fig.~\ref{fig:1_teaser}-left). On the more challenging PhlatCam dataset, NSDD continues to generate plausible reconstructions with natural structure and texture, despite differences in acquisition conditions. Additional qualitative examples in the supplementary material illustrate failure cases (e.g., oversmoothing or brightness artifacts), which we attribute to limitations of the forward model and pseudo-inverse.

We do not report quantitative results on the PhlatCam, and instead use it to provide qualitative evidence of generalization across acquisition conditions. The qualitative results suggest that the proposed distillation strategy generalizes beyond the face-centered Lensless-FFHQ dataset.

Overall, these results demonstrate that distilling structured diffusion-based inference enables fast reconstruction while retaining high measurement consistency and perceptual quality.

\begin{figure*}
    \centering
    \includegraphics[width=0.95\textwidth]{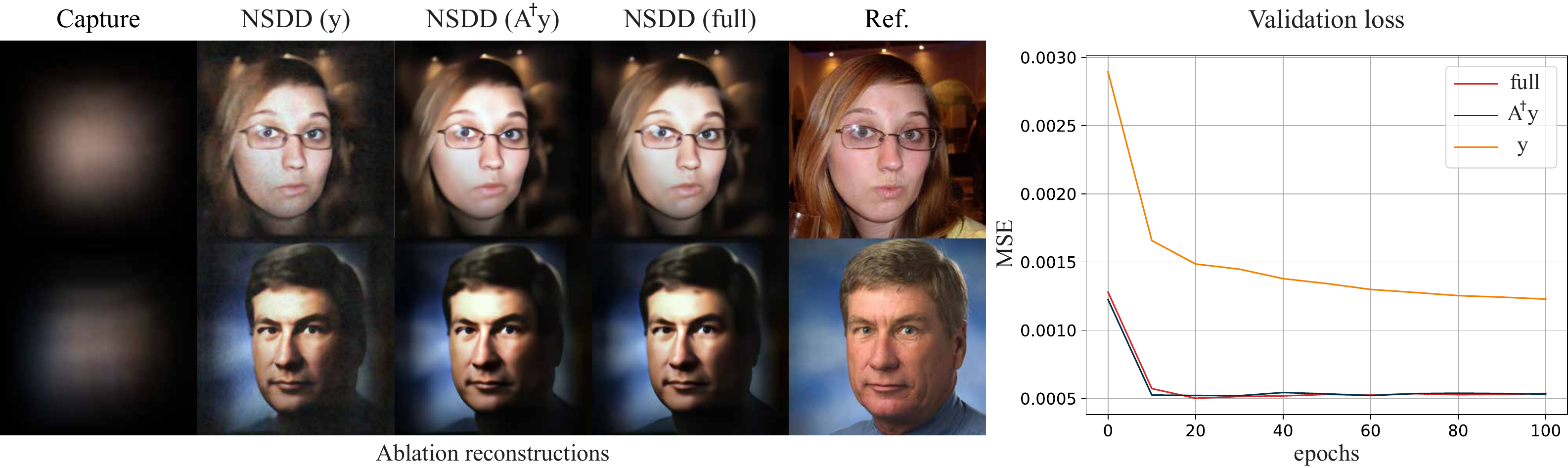}
    \caption{Ablation on conditioning signals in NSDD.
Left: Reconstructions using NSDD conditioned on $\mathbf{y}$ only, $A^\dagger \mathbf{y}$ only, and both signals (full), compared to the reference image. Conditioning on $A^\dagger \mathbf{y}$ preserves structural and perceptual quality but yields slightly worse measurement consistency, while conditioning on $\mathbf{y}$ alone degrades reconstruction quality. 
Right: Validation loss curves for the three variants. Using $\mathbf{y}$ alone leads to slower convergence and higher loss, indicating increased training difficulty, whereas conditioning on $A^\dagger \mathbf{y}$ stabilizes training.}
    \label{fig:ablation}
\end{figure*}

\subsection{Ablation on conditioning signals}
\label{sec:nsdd_ablation}
\red{Next}, we ablate the role of the conditioning signals in NSDD by training variants using only $\mathbf{y}$ or only $A^\dagger \mathbf{y}$, and comparing them to the fully conditioned model. Fig.~\ref{fig:ablation}-(left) shows that conditioning on $\mathbf{y}$ alone produces noisy reconstructions, while using $A^\dagger \mathbf{y}$ alone achieves perceptual quality comparable to the full model. This behavior is explained by the training dynamics in Fig.~\ref{fig:ablation}-(right), where the $\mathbf{y}$-only variant exhibits higher validation loss, indicating increased optimization difficulty. Quantitative results (Tab.~\ref{tab:ablation}) further show that, despite similar perceptual and structural metrics, conditioning on $A^\dagger \mathbf{y}$ alone leads to worse measurement consistency compared to the full model, due to the absence of direct measurement information.

\begin{table}[t]
\centering
\caption{Ablation on conditioning signals. Comparison between NSDD conditioned on $\mathbf{y}$, $A^\dagger \mathbf{y}$, and both (full). \textbf{Bold} represents the best.}
\label{tab:ablation}
\setlength{\tabcolsep}{4pt}
\begin{tabular}{lcccc}
\toprule
Method & Consist. ($10^{-4}$) $\downarrow$ & SSIM $\uparrow$ & LPIPS $\downarrow$ & CLIP-IQA $\uparrow$ \\
\midrule
NSDD ($\mathbf{y}$) & 0.7008 & \textbf{0.5501} & 0.4240 & 0.4442\\
NSDD ($A^\dagger \mathbf{y}$) & 0.1680 & 0.4998 & \textbf{0.3790} & 0.4923\\
NSDD (full) & \textbf{0.1156} & 0.4984 & 0.3807 & \textbf{0.4954}\\
\bottomrule
\end{tabular}
\end{table}


\begin{table}[t]
\centering
\caption{Relationship between number of diffusion sampling steps and inference speed, measurement consistency and perceptual quality for DDNM+.}
\label{tab:ddnm_ablation}
\setlength{\tabcolsep}{4pt}
\begin{tabular}{lccc}
\toprule
Method & TIME (s) $\downarrow$ & Consist. ($10^{-4}$) $\downarrow$ & LPIPS $\downarrow$  \\
\midrule
DDNM+ (2 steps)     & 0.0615 & 0.1245 & 0.5882 \\
DDNM+ (10 steps)     & 0.1900 & 0.0969 & 0.4029  \\
DDNM+ (50 steps)     & 0.9750 & 0.0973 & 0.3960  \\
DDNM+ (1000 steps)  & 20.8908 & 0.0928 & 0.3717  \\
\bottomrule
\end{tabular}
\end{table}

\subsection{Impact of number of DDNM+ sampling steps}
\label{sec:ddnm_ablation}

A natural approach to improving the time efficiency of diffusion-prior iterative methods is to reduce the number of sampling steps. We can evaluate this trade-off by varying the number of DDNM+ sampling steps between the original $1{,}000$ steps and the aggressively reduced $2$ steps and measuring metrics that capture perceptual quality (i.e., LPIPS), measurement consistency, and inference time.

Results are reported in Tab.~\ref{tab:ddnm_ablation}. Reducing the number of sampling steps leads to a degradation in both perceptual quality and consistency, with a pronounced drop at very low step counts (e.g., $2$ steps). While inference time decreases significantly, even aggressively reduced configurations remain slower than NSDD, while achieving substantially worse reconstruction quality.

These results highlight the inherent limitation of iterative diffusion-based methods: reducing computational cost comes at the expense of both fidelity and perceptual realism, motivating the need for alternatives such as NSDD.

\begin{figure}
    \centering
    \includegraphics[width=\linewidth]{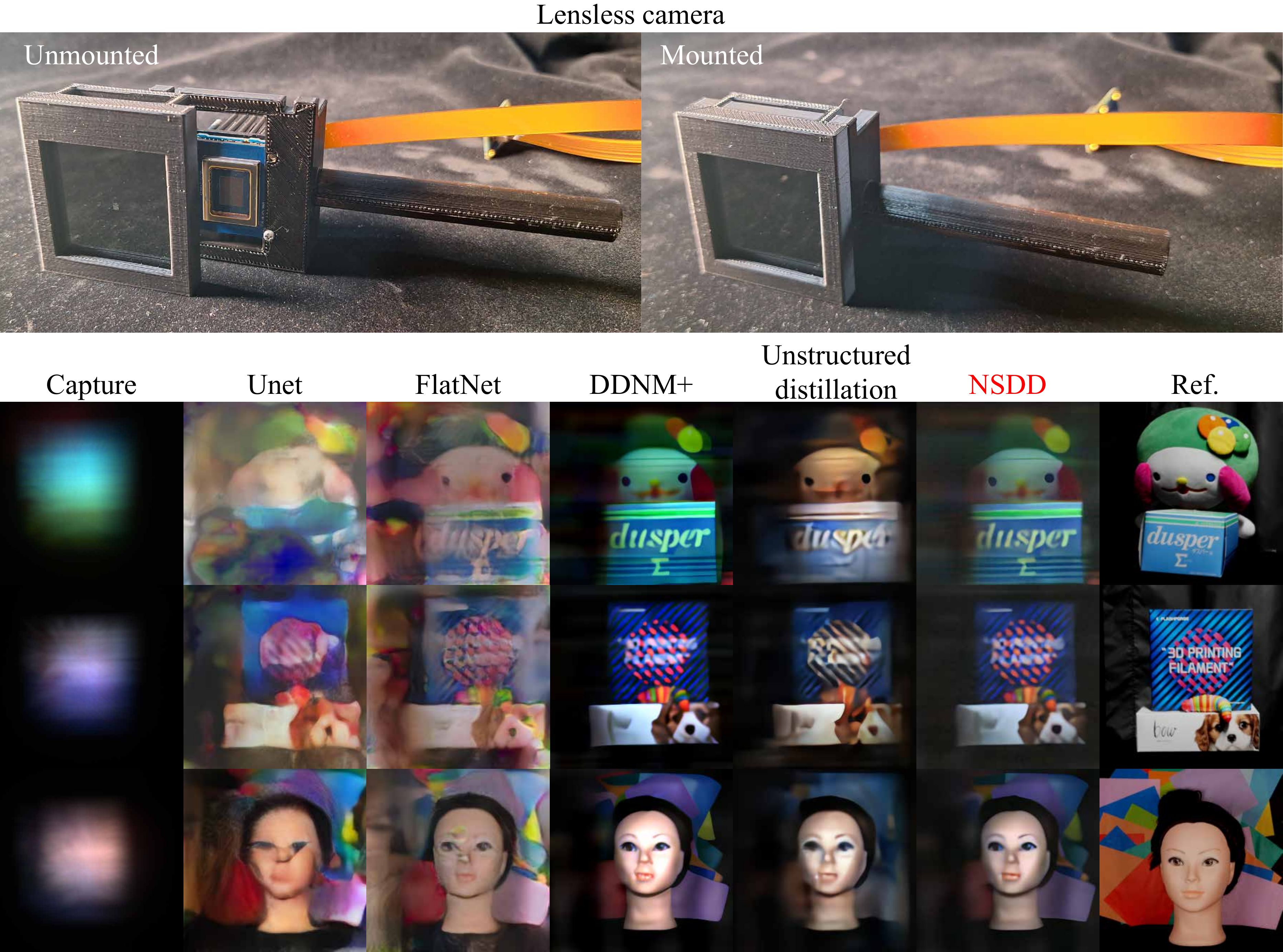}
    \caption{Reconstructions of real scenes using our prototype lensless camera with models trained only on monitor-displayed face images.}
    \label{fig:ood_experiment}
\end{figure}

\subsection{Experiment with real objects}
\label{sec:real_objects}
\red{The lensless captures from Lensless-FFHQ and PhlatCam datasets were taken using images displayed from monitors, and so it brings the question on how well each of the trained models generalize to real scenes. To answer this question, we captured new real out-of-distribution scenes with the Lensless-FFHQ camera and reconstructed them using the models trained only on the Lensless-FFHQ dataset (see Fig.~\ref{fig:ood_experiment}). Unlike the Lensless-FFHQ dataset, which contains monitor-captured faces, these scenes include both real face-like and non-face physical objects. Fig.~\ref{fig:ood_experiment} shows that supervised methods fail, whereas DDNM+ and NSDD produce high-quality reconstructions in both cases. This supports the range-null decomposition used by DDNM and NSDD as a strong framework for practical lensless imaging. Reference images (i.e., Ref.) are taken with a lensed camera.}

\begin{table}[t]
\centering
\caption{\textbf{Ablation on distillation strategy.}
Comparison between unstructured full-reconstruction distillation and the proposed NSDD null-space distillation.}
\label{tab:nsdd_distillation_ablation}
\setlength{\tabcolsep}{4pt}
\begin{tabular}{lccc}
\toprule
Method & Consist. ($10^{-4}$) $\downarrow$ & SSIM $\uparrow$ & LPIPS $\downarrow$ \\
\midrule
Unstructured distillation  & \textbf{0.1061} & 0.4719 & 0.4071   \\
NSDD                & 0.1156 & \textbf{0.4984} & \textbf{0.3807} \\
\bottomrule
\end{tabular}
\end{table}

\subsection{Ablation on null-space distillation choice}

\red{To further motivate our DDNM$+$ null-space distillation design, we perform an ablation in which we distill a single-pass full-reconstruction model instead of the proposed range-null decomposition. We refer to this baseline as unstructured distillation, since it does not explicitly decompose the reconstruction process into range- and null-space components. This setting is conceptually closer to general diffusion-prior reconstruction methods such as DPS and DAPS, which optimize or sample the full image under a measurement-consistency objective, rather than explicitly separating the measurement-consistent component from the learned prior correction. As in NSDD, the unstructured model is trained using DDNM$+$ teacher reconstructions as targets. However, instead of predicting only the null-space correction, it directly maps the lensless measurement to the full reconstruction. For a fair comparison, we use the same architecture as NSDD, consisting of an input-reducer U-Net followed by the pretrained diffusion U-Net, with the input-reducer adapted to three input channels instead of six.}

\red{The results on Lensless-FFHQ are shown in Tab.~\ref{tab:nsdd_distillation_ablation}. Compared with unstructured distillation, NSDD achieves better perceptual and structural reconstruction quality, indicating that explicitly structuring the distillation around the range-space component leads to more faithful image recovery. The unstructured baseline obtains slightly better measurement-consistency scores; however, this comes at the cost of degraded perceptual and structural quality. We further evaluate the unstructured model on real out-of-distribution scenes in Fig.~\ref{fig:ood_experiment}. Although it remains more effective than fully supervised baselines such as FlatNet and U-Net, it exhibits a clear loss of detail and color fidelity compared with both the DDNM$+$ teacher and NSDD. These results show that directly distilling full reconstructions does not preserve the teacher's robustness as effectively as the proposed null-space formulation. Overall, this ablation supports our null-space distillation design while also suggesting that diffusion-model distillation is a promising direction for future lensless imaging research.}

\subsection{Time requirements for NSDD}

\red{As a final consideration, we report the computational time required by the NSDD training pipeline. NSDD is ground-truth-free and is therefore closer in spirit to unsupervised methods such as ADMM and DDNM$+$. However, unlike purely optimization-based methods, it requires an offline distillation stage based on teacher reconstructions. We therefore report the time required for the two pre-inference stages: (1) generating DDNM$+$ teacher targets and (2) training the student network.}

\red{\textbf{Teacher-target generation.}
Teacher targets are generated using the DDNM$+$ implementation. Since DDNM$+$ does not require gradient computations and uses efficient Wiener deconvolution as the pseudo-inverse, it has lower memory requirements and faster inference than gradient-based diffusion methods such as DPS, as depicted in Tab.~\ref{tab:restoration_metrics}. Using an RTX 5090 GPU, we reconstruct 48 lensless measurements per batch in $431.96$ seconds on average, including I/O operations. For the $21,000$ captures in Lensless-FFHQ, this corresponds to approximately 52 hours on a single GPU.}

\red{\textbf{Model training.}
After generating the DDNM$+$ teacher targets, we train the NSDD reconstruction network using the Lensless-FFHQ split described in Sec.~\ref{sec:experimental_setup}. Training takes $921.38$ seconds per epoch on average, for a total training time of approximately 25 hours on an NVIDIA A100 GPU.}

\red{\textbf{Discussion.}
Compared with purely unsupervised methods such as ADMM and DDNM$+$, NSDD introduces an offline training overhead. However, this cost is incurred only once for a given imaging setup. After training, NSDD enables fast feed-forward inference while maintaining strong reconstruction quality and measurement consistency, as shown in Tab.~\ref{tab:restoration_metrics}. Moreover, its generalization to unseen real captures, including out-of-distribution objects as shown in Sec.~\ref{sec:real_objects}, suggests that the trained model is robust to practical input variations and does not need to be retrained for every new scene or object category.}

\section{Conclusion}
\label{sec:conclusion}

In this work, we analyzed reconstruction strategies for lensless imaging and identified a fundamental trade-off between measurement consistency, perceptual quality, and inference speed, showing that traditional, supervised, and diffusion-prior methods occupy distinct regions in this space. We further showed that diffusion-prior iterative methods can achieve high perceptual quality while maintaining measurement consistency, but at the cost of slow iterative sampling.

Motivated by this analysis, we proposed \emph{Null-Space Diffusion Distillation} (NSDD), a reconstruction model that distills the structured range-null decomposed behavior of a DDNM$+$ solver into an efficient single-pass feed-forward neural network. 
In doing so, NSDD retains the benefits of diffusion-based inference while eliminating its iterative computational burden and thus \red{significantly speeding up the reconstruction process}. 

Experimental results demonstrate that NSDD bridges the gap between traditional, supervised, and diffusion-based approaches, enabling high perceptual reconstruction quality and measurement fidelity with fast inference speed. Beyond this specific setting, our results show that the structured behavior of diffusion-based inference—particularly when grounded in the forward model—can be distilled into efficient predictors without sacrificing key properties such as physical consistency and perceptual realism. \red{Additionally, we showed that NSDD generalizes to effectively out-of-distribution scenes.}

\noindent\textbf{Future directions.}
The experiments in this work assume a shift-invariant point spread function (PSF), enabling a convolutional forward model and a Wiener filter approximation of the pseudo-inverse operator $A^\dagger$. While this simplification is common, it limits the fidelity of the range-space reconstruction used as the anchor in our framework. Improving the accuracy of the forward model and its corresponding pseudo-inverse—e.g., by accounting for spatially varying PSFs~\cite{Kuo_etal_localConv_2020_OptExp, Cai_etal_phocolens_2024_NeurIPS} or more accurate inversion procedures—would provide a more informative range-space estimate $A^\dagger \textbf{y}$. This, in turn, could allow diffusion-prior methods such as DDNM+ and our proposed NSDD to more effectively recover fine details, leading to improved reconstruction quality.

\ifpeerreview \else
\section*{Acknowledgments}
This work was supported by JST FOREST (Grant JPMJFR206K) and Shimadzu Research Grants.
\fi

\bibliographystyle{IEEEtran}
\bibliography{references}

\ifpeerreview \else


\begin{IEEEbiographynophoto}{Jose Reinaldo Cunha Santos A V Silva Neto}
received the B.S. degree in Control and Automation Engineering and the M.S. degree in Computer Science from the University of Brasília, Brazil, in 2019 and 2021, respectively, and the Ph.D. degree in Computer Science from the University of Osaka, Japan, in 2026. He is currently a Postdoctoral Researcher with the Intelligent Sensing Laboratory, the University of Osaka. His research interests include computational photography, lensless imaging, and deep learning for computer vision. He is a member of the Optical Society of Japan and was a recipient of the Student Presentation Award at the 13th Japan–Korea Workshop on Digital Holography and Information Photonics (DHIP 2023).
\end{IEEEbiographynophoto}


\begin{IEEEbiographynophoto}{Hodaka Kawachi}
 received the B.S. in engineering and M.S. in computer science from Osaka University, in 2022 and 2024 respectively. Currently, he is a PhD candidate on the computer science department of Osaka University. His research interests include computational photography, with a preference for depth imaging, and deep learning techniques applied to computer vision.
 He has been awarded several honors, including the 27th Meeting on Image Recognition and Understanding (MIRU2024) MIRU Student Award (2024), the Graduate School of Information Science and Technology Award from Osaka University (2023), the 13th Japan-Korea Workshop on Digital Holography and Information Photonics (DHIP2023) Student Presentation Award (2023), the 25th Meeting on Image Recognition and Understanding (MIRU2022) MIRU Student Award (2022), and the Konica Minolta Optics Future Encouragement Award (2024). He has also been selected as a recipient of the JSPS DC1 fellowship since 2024.
\end{IEEEbiographynophoto}

\begin{IEEEbiographynophoto}{Yasushi Yagi}
(Senior Member, IEEE) received the Ph.D. degree from Osaka University, in 1991. In 1985, he joined the Product Development Laboratory, Mitsubishi Electric Corporation, where he was involved in robotics and inspections. He became a Research Associate, in 1990, a Lecturer, in 1993, an Associate Professor, in 1996, and a Professor, in 2003, with Osaka University, where he was the Director of SANKEN (The Institute of Scientific and Industrial Research), from 2012 to 2015. He was the Executive Vice President of Osaka University, from 2015 to 2019. His research interests include computer vision, pattern recognition, biometrics, human sensing, medical engineering, and robotics. He is a fellow of the IPSJ and a member of the IEICE and the RSJ. He is a member of the Editorial Board of the {\it International Journal of Computer Vision}. He is the Vice President of the Asian Federation of Computer Vision Societies. He was awarded the ACM VRST2003 Honorable Mention Award, the IEEE ROBIO2006 Finalist of the T. J. Tan Best Paper in Robotics, the IEEE ICRA2008 Finalist for the Best Vision Paper, the PSIVT2010 Best Paper Award, the MIRU2008 Nagao Award, the IEEE ICCP2013 Honorable Mention Award, the MVA2013 Best Poster Award, the IWBF2014 IAPR Best Paper Award, and the {\it IPSJ Transactions on Computer Vision and Applications} Outstanding Paper Award (2011 and 2013). International conferences for which he has served as the Chair include ROBIO2006 (PC), ACCV (2007PC and 2009GC), PSVIT2009 (FC), and ACPR (2011PC, 2013GC, 2021GC, and 2023GC). He has al
so served as an Editor for the IEEE ICRA Conference Editorial Board (2008 and 2011). He was the Editor-in-Chief of the {\it IPSJ Transactions on Computer Vision and Applications}.
\end{IEEEbiographynophoto}

\begin{IEEEbiographynophoto}{Tomoya Nakamura}
is an Associate Professor at the Graduate School of Engineering Science, The University of Osaka (formerly Osaka University), working on computational imaging and holography. He received his Ph.D. from The University of Osaka in 2015. From 2015 to 2020, he served as an Assistant Professor at Tokyo Institute of Technology. He is a member of Optica. His honors include the Konica Minolta Imaging Science Award (2018), the 1st Kyowa Technologies ICT Encouragement Award (2025), and the Young Scientists’ Award from MEXT (Ministry of Education, Culture, Sports, Science and Technology), Japan (2026).
\end{IEEEbiographynophoto}


\fi

\clearpage
\appendices

\setcounter{section}{0}
\setcounter{figure}{0}
\setcounter{table}{0}
\setcounter{equation}{0}

\renewcommand{\thesection}{S\arabic{section}}
\renewcommand{\thefigure}{S\arabic{figure}}
\renewcommand{\thetable}{S\arabic{table}}
\renewcommand{\theequation}{S\arabic{equation}}

\supplementtitle


\begin{figure}
    \centering
    \includegraphics[width=\linewidth]{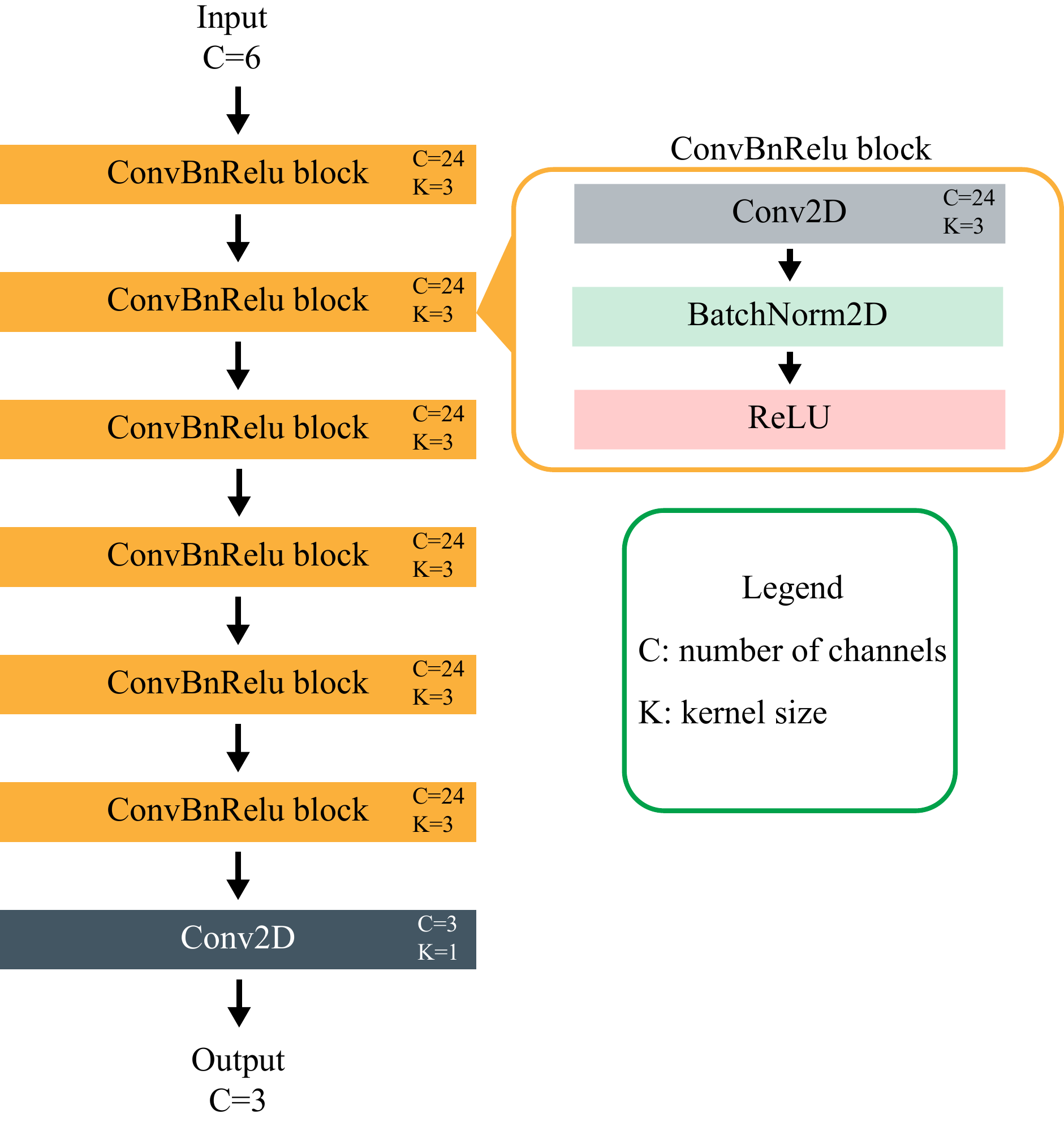}
    \caption{Structure of the input reducer network.}
    \label{fig:supp_small_net}
\end{figure}

\section{Additional details about implementation of diffusion-prior methods for lensless imaging}
\label{sec:diff_prior_details}
\vspace{0.5em} \noindent{\bf Data preprocessing.} For both datasets (i.e., lensless FFHQ and PhlatCam), we preprocess the lensless measurements and PSF by first cropping them from the original resolution to a square size, and subsequently resizing them to a resolution so that the reconstructed scene is constrained to the central $256\times 256$ pixels, to make them compatible with the resolution for which the diffusion models were pretrained for. For the lensless FFHQ dataset, from the original capture size of $1464\times 1936$ we crop a region of $1200\times 1200$ and downsample to $512\times 512$. For the PhlatCam dataset, we crop from the original resolution of $1518\times 2012$ to $1518\times 1518$ and downsample to $950\times 950$. 

\vspace{0.5em}
\noindent\textbf{Resolution mismatch between diffusion and camera.}
Lensless reconstructions exhibit dark, non-imaged borders, so systems are designed such that recoverable content lies near the sensor center. In contrast, our pretrained diffusion prior expects full-frame lensed images without dark margins. We bridge the gap between these domains with an invertible \emph{center-crop / zero-pad} pair.

Let $C_{256}:\mathbb{R}^{H\times W\times3}\!\to\!\mathbb{R}^{256\times256\times3}$ extract the centered $256{\times}256$ crop, and let
$P_{H,W}:\mathbb{R}^{256\times256\times3}\!\to\!\mathbb{R}^{H\times W\times3}$ pad a $256{\times}256$ image into the $H{\times}W$ sensor grid by zero-padding around the crop. We use $C_{256}$ whenever a padded image is to be used by the diffusion model (e.g., projections onto range- and null-spaces by DDNM), and $P_{H,W}$ whenever an image output by the diffusion model is to be passed to a lensless operator (forward model, Wiener deconvolution). Dataset-specific sizes are $H{=}W{=}950$ for \textit{PhlatCam} and $H{=}W{=}512$ for \textit{Lensless FFHQ}. This mapping satisfies $C_{256}\!\circ\!P_{H,W}=\mathrm{I}$ and restores the expected dark border for physical constraints.

\section{Additional details about preprocessing, alignment, and normalization used by all methods}
\vspace{0.5em}
\noindent\textbf{Measurement preprocessing.} This process is done as presented in the previous section. 

\vspace{0.5em}
\noindent\textbf{Reconstruction alignment.} Since lensless reconstructions from traditional and diffusion-prior methods and reference images are not perfectly aligned in practical lensless systems, we further \textit{align} each reconstruction to its corresponding reference image using an affine transformation estimated via feature matching and RANSAC. Not only pixel-based metrics such as PSNR and SSIM, but all restoration metrics are computed on the aligned images restricted to the valid overlapping region, to ensure fairness.

\vspace{0.5em}
\noindent\textbf{Reconstruction normalization.} Finally, all reconstructed images are \textit{normalized} to the range $[0,1]$ prior to computing evaluation metrics to ensure compatibility with the reference images. To avoid biasing the measurement consistency metric due to normalization, we estimate per-image scaling and offset parameters ($a$ and $b$) that minimize the forward-model residual as $\|\boldsymbol{y} - \boldsymbol{A}(a\hat{\boldsymbol{x}}+b)\|_2$.

\section{Datasets and train-test splits}
We use two datasets for the experiments in this paper. One is the PhlatCam from~\cite{Khan_etal_flatnet_PAMI_2022}, and the other is the lensless FFHQ from~\cite{Neto_etal_SelfNeuralLensless_2025_OptReview}. For the lensless FFHQ reconstructions, we use the pretrained diffusion model from~\cite{Chung_etal_DPS_2023_ICLR}. Their pre-training of the diffusion model held out the first $1,000$ images from the original FFHQ dataset, so we follow the same approach and keep the first $1,000$ images on the test set when training our NSDD network. As for PhlatCam dataset, we use the pre-trained diffusion model trained on the ImageNet dataset by~\cite{dhariwal_nichol_guideddiff_NIPS_2021}. The PhlatCam (general restricted-aperture coded mask camera) dataset used by the FlatNet authors~\cite{Khan_etal_flatnet_PAMI_2022} consists of a smaller number of 2592 lensless captures\footnote{https://siddiquesalman.github.io/flatnet/}. Because of that, we manually selected a smaller number (i.e., 10) of them to hold on the testing set. The files that we used on the testing set are those described by the numberings: n01644373\_12084, n02168699\_7495, n03376595\_294, n03445777\_10782, n03785016\_15679, n03793489\_8764, n04584207\_17205, n04591713\_1640, n07720875\_706, n07720875\_2738. Our objective was to select samples for images that the content could be easily identified by the reader (e.g., natural images of insects, person, and common objects).

\begin{figure}
    \centering
    \includegraphics[width=\linewidth]{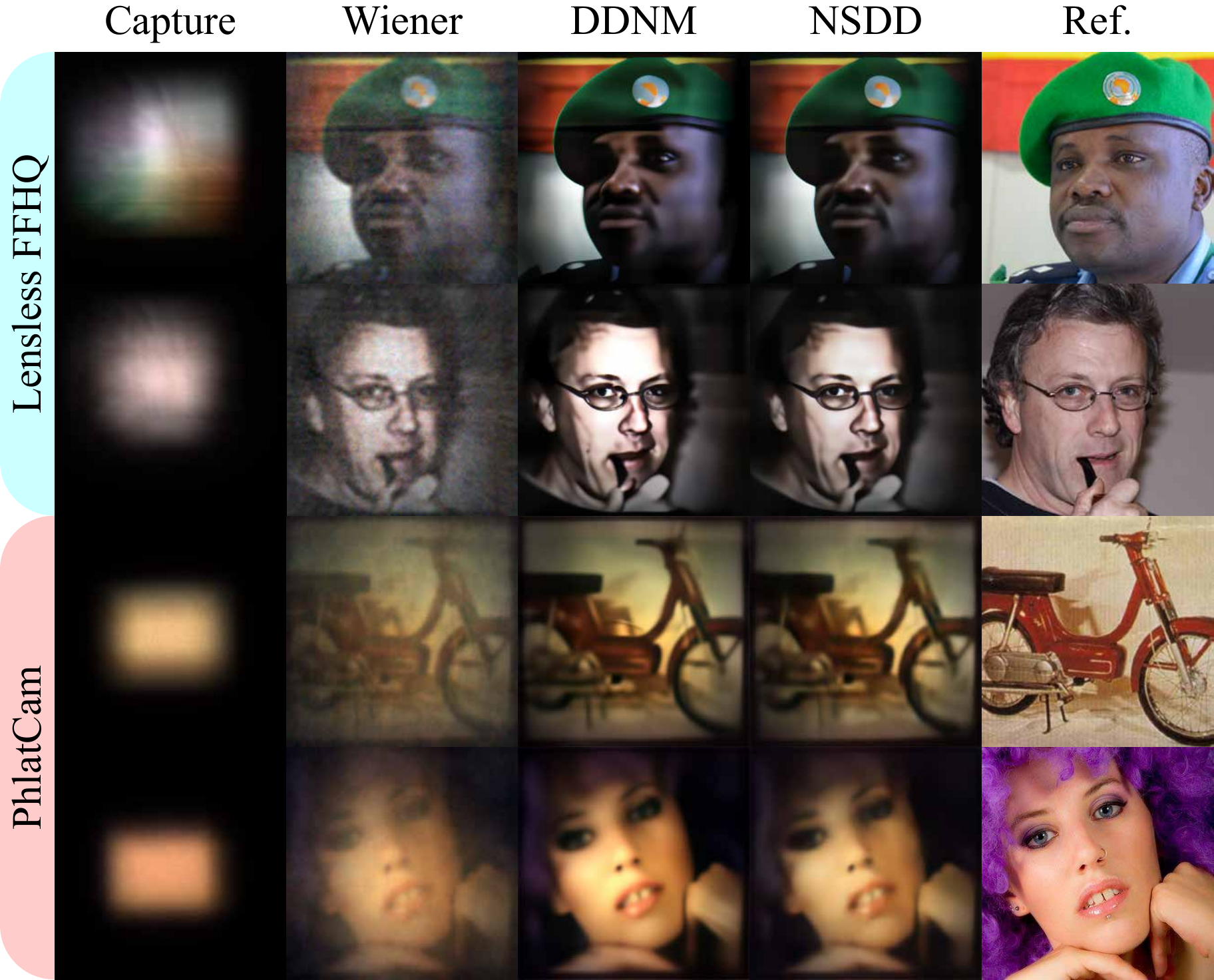}
    \caption{Reconstructions with lower quality for both lensless FFHQ and PhlatCam datasets.}
    \label{fig:supp_bad_recon}
\end{figure}

\section{Neural network architectures for NSDD}
The architecture for the student network is composed of two neural models arranged in sequence. The first is the smaller input reducer network, responsible to reduce the number of channels of the input images from 6 down to 3. The second is the denoising UNet used by the pretrained diffusion model, that we initialize with the trained weights and further optimize in an end-to-end scenario jointly with the input reducer network. The structure of the input reducer network is shown in Fig.~\ref{fig:supp_small_net}.

\section{Lower quality reconstruction samples for DDNM+ and NSDD}
Despite the high-quality reconstructions achieved by the results shown throughout the paper, there are some settings that make both DDNM$+$ and NSDD methods produce reconstructions with lower quality. Some examples are shown in Fig.~\ref{fig:supp_bad_recon}, where oversmoothing of high-frequency information and unnatural brightness is generated. We argue that these are not a limitation of the diffusion-prior method itself, but rather of the lensless imaging model used. More specifically, if the approximations for the lensless forward model (i.e., shift-invariant 2D convolution) and for the pseudo-inverse (i.e., Wiener deconvolution), can be further improved, then better range-anchor information can be retrieved from the lensless capture adn that would allow the diffusion-prior to fill the null-space and denoise it more effectively.

\section{Parameter-sweep analysis for diffusion-prior methods}
\label{sec:diffpriors}


\begin{figure*}
    \centering
    \includegraphics[width=0.9\linewidth]{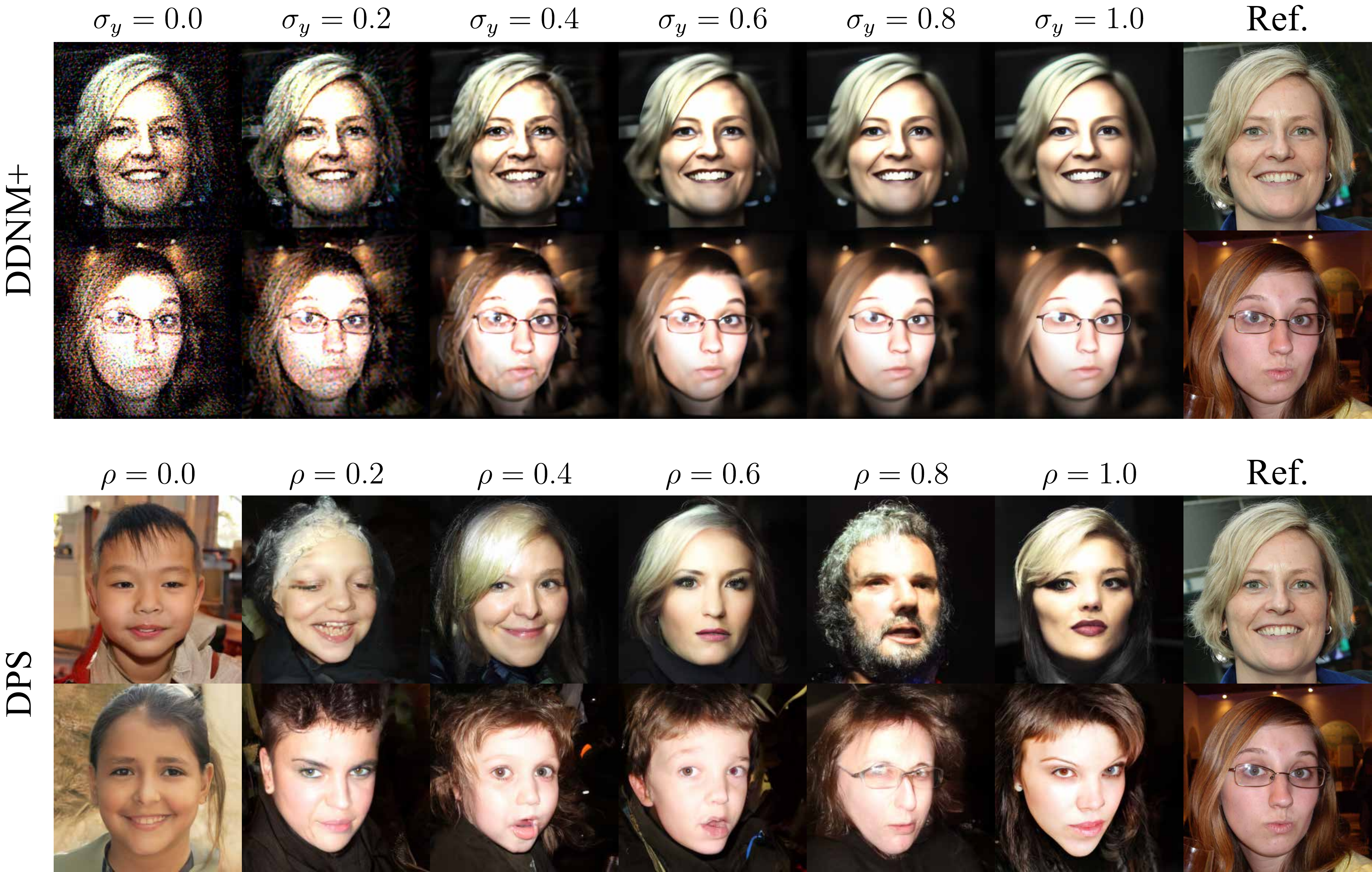}
    \caption{Parameter sweep for DPS and DDNM$+$ algorithms for samples of the lensless FFHQ dataset.}
    \label{fig:dps_vs_ddnm}
\end{figure*}
\begin{figure}
    \centering
    \includegraphics[width=0.8\linewidth]{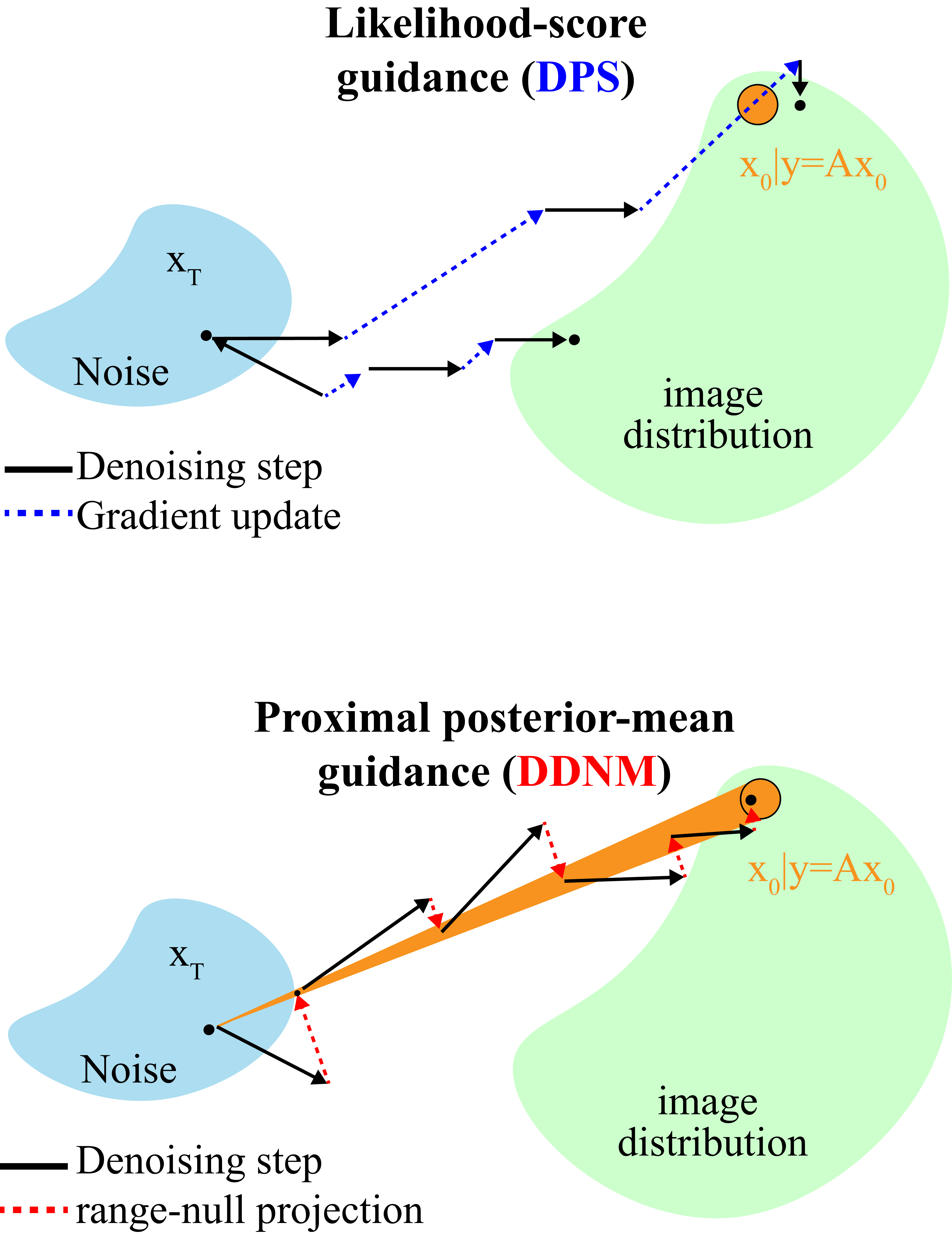}
    \caption{Depiction of the two types of diffusion guidance approaches analyzed on this paper. On the top, the likelihood-score guidance  with an under-steered trajectory (small gradient updates) and an over-steered trajectory (large gradient updates) for the same sample, showcasing the challenging tradeoff between consistency and quality. On the bottom, the proximal posterior-mean guidance approach enforces a projection onto a consistent path that leads to plausible reconstructions on the image domain.}
    \label{fig:diff_prior_diagram}
\end{figure}

While they have been thoroughly tested for several inverse imaging tasks like image deblurring, inpainting and super resolution, diffusion-prior techniques in the literature have foregone the lensless imaging task. In this section we demonstrate that techniques developed in the literature do not necessarily translate to lensless imaging deconvolution. We do so by comparing two seminal works (i.e., DPS and DDNM$+$) applied to the lensless imaging scenario. We use the same setup used in Sec.~3, with the lensless-FFHQ dataset.

\vspace{0.5em} \noindent{\bf Experiment setup} consists of a parameter sweep to analyze the impact of parameter fine-tuning on the reconstructions from DPS and DDNM$+$ techniques. More specifically, we vary the $\rho$ (DPS) and $\sigma_y$ (DDNM$+$) in equally spaced intervals inside the range $[0, 1]$ to evaluate the impact of hyperparameter fine-tuning on reconstruction quality and consistency. Similar to the official implementation of DPS, we retain a constant coefficient $\rho$ throughout optimization. The pretrained diffusion model is sampled for $1,000$ steps for both DPS and DDNM$+$ methods. The reconstructions are shown in Fig.~\ref{fig:dps_vs_ddnm}. Similar to the diagram in Fig.~\ref{fig:diff_prior_diagram}-(top), we can see that the under-steering of the guidance term for small values of $\rho$ produce low-consistency images, while high values over-steer the denoised image into closer but still not consistent regions. DDNM$+$, on the other hand, produces results compatible with the representation in Fig.~\ref{fig:diff_prior_diagram}-(bottom), always achieving plausible reconstructions, but with high values of $\sigma_y$ producing smoothed images and low values producing noisy images since the range space is not denoised.

From the observed results, we claim that despite its success in several inverse imaging tasks, DPS fails in reconstructing both consistent and high-quality images due to the nature of the noisy and highly multiplexed lensless measurements. DDNM$+$, however, achieves significant success in reconstructions from a perceptual quality standpoint, while also having faster inference time.

\end{document}